\documentclass[conference,letterpaper,twoside,final,10pt]{ieeeconf}
\usepackage{graphicx}
\usepackage{amssymb}
\usepackage{amsmath}
\usepackage{cite}
\usepackage{comment}
\usepackage{color}
\usepackage{url}
\usepackage{alltt}
\usepackage{cleveref}
\usepackage{tikz}
\usepackage{pgfplots}
\usepackage{pgf-umlsd}
\usepackage{multirow}
\usepackage{algorithm}
\usepackage[noend]{algpseudocode}
\usepackage{bigfoot}
\usepackage{paralist}
\usepackage{tabularx}
\usepackage[whole]{bxcjkjatype}
\usepackage{ifthen}
\usepackage{siunitx}
\usepackage{threeparttable}
\usepackage[hang,small,bf]{caption}
\usepackage[subrefformat=parens]{subcaption}
\pgfplotsset{compat=1.14}
\IEEEoverridecommandlockouts

\graphicspath{ {./} }

\title{\LARGE \bf
 Exploration of Reinforcement Learning for Event Camera using Car-like Robots 
}

\author{
Riku Arakawa$^{*}$$^{\dagger}$,
Shintaro Shiba $^{**}$$^{\dagger}$,
\thanks{$^{*}$ The University of Tokyo. \newline \url{arakawa-riku428@g.ecc.u-tokyo.ac.jp}}
\thanks{$^{**}$ Keio University. \newline 
\url{sshiba@keio.jp}}
\thanks{$^{\dagger}$ equal contribution (ordered alphabetically) }}

\begin{document}
\maketitle

\begin{abstract}
We demonstrate the first reinforcement-learning application for robots equipped with an event camera.
Because of the considerably lower latency of the event camera, it is possible to achieve much faster control of robots compared with the existing vision-based reinforcement-learning applications using standard cameras.
To handle a stream of events for reinforcement learning, we introduced an image-like feature and demonstrated the feasibility of training an agent in a simulator for two tasks: fast collision avoidance and obstacle tracking.
Finally, we set up a robot with an event camera in the real world and then transferred the agent trained in the simulator, resulting in successful fast avoidance of randomly thrown objects.
Incorporating event camera into reinforcement learning opens new possibilities for various robotics applications that require swift control, such as autonomous vehicles and drones, through end-to-end learning approaches.
\end{abstract}

\section{Introduction} % Riku

Robot automation has been prevailing in our society to assist us in various industries \cite{bahrin2016industry, wollschlaeger2017future}.
Reinforcement learning is a highly promising technique that enables robots to acquire skills without the human effort of manually designing rules for possible input--output controlling patterns \cite{gullapalli1994acquiring}.
Furthermore, transfer learning \cite{pan2009survey} can release us from training robots in the real world, which is often very costly.
Instead, we can train them in a prepared simulator as long as its environment is similar to the real-world setting.

So far, most of the current reinforcement learning systems assume signals from standard RGB-cameras as input.
This is attributed to the development of neural networks and computing power that allow us to handle high-dimensional data with practical speed to get rich information for robots.
However, relying on image inputs restricts the robot's control frequency to at most the sampling frequency of the image sensor.

In fact, most of the standard image sensors take images at $30$--$60$~Hz (frames per second).
This limitation can be a serious problem where high-speed control is required, such as in autonomous vehicles or drones because if a vehicle moves at 30~m/s (67~mi/h), it moves one meter in 33~ms, which is ``blind time'' without any signal between each frame at 30 Hz.
In addition, the rotational movement of robots results in the rapid motion of surrounding objects, sometimes with motion blur, which is a critical problem in robots such as autonomous drones.

Recently, as another ``eye'' for robots, event cameras have emerged \cite{posch2014retinomorphic}.
The event camera, or event-based camera, is a neuromorphic vision sensor.
It enables us to process stream data at the sub-millisecond resolution, which is much faster than the processing speed of standard CMOS or CCD cameras. 
By using event cameras, the frequency of controlling robots is no longer limited to 30 Hz or 60 Hz.
Moreover, unlike standard high-speed cameras, it is small enough, and its power consumption is low enough to attach to mobile robots.
Therefore, the present robotics community has many expectations for its applications.

\begin{figure}[t]
\centering
\includegraphics[width=0.49\textwidth]{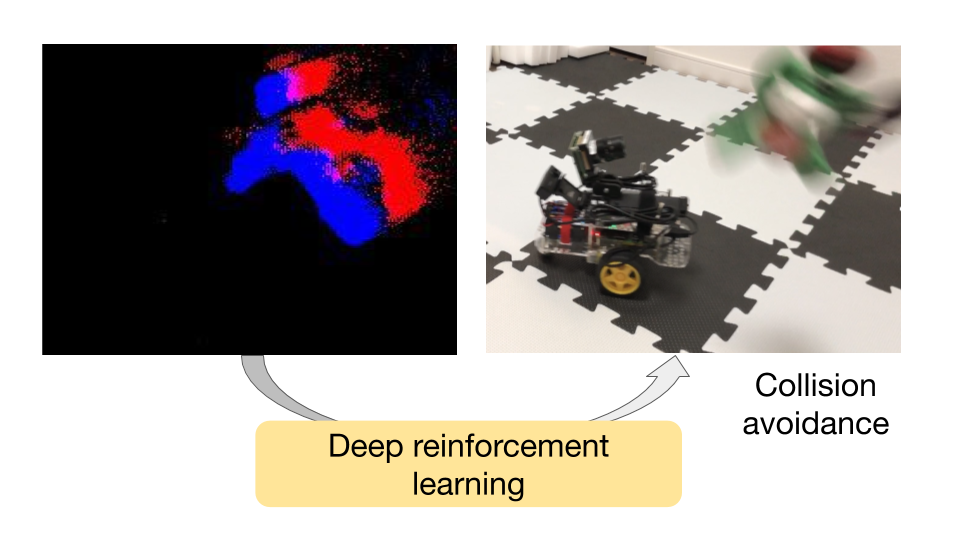}
\caption{The agent trained in the simulator can be transferred to a real robotic car to perform fast collision avoidance. The left image is an image-like feature of events, which is an input for deep reinforcement-learning agent. The right photo is the real robotic car equipped with an event camera (DAVIS240)\cite{brandli2014240}. For details, please refer to the video https://youtu.be/xc2nKJ1TLTY.}
\label{fig:eye-catch}
\end{figure}

In this paper, we demonstrate a real-world reinforcement-learning application with an event camera, the policy of which is obtained in a simulation environment.
First, we present an efficient method to directly create image-like features of simulated events for input to conventional reinforcement learning algorithms in a simulator.
Then, we show that we can successfully train an agent in the simulator to learn collision avoidance and tracking objects, both of which require fast control.
Finally, we demonstrate that the trained agent was able to be transferred into a real robot with an attached event camera by converting a stream of events into image-like features (Fig. \ref{fig:eye-catch}).

To the best of the authors' knowledge, this is the first study to merge reinforcement learning with the event camera and launch a demonstration of the trained agent.
Along with the empirical result, our exploratory implementation will open a new path for the vision-based automated robots with very low latency.

\section{Literature}
\subsection{Event Camera Overview} 

Standard cameras record scenes at fixed time intervals and output a sequence of image frames.
In contrast, an event camera outputs a stream of asynchronous events at microsecond resolution, indicating when individual pixels record the log intensity to change over a preset threshold size.
While standard cameras have blind time-intervals and may also send temporally redundant data if the captured scene does not change between frames, event cameras output changes at more precise times with less redundancy \cite{gallego2019eventbased}.
Hence, event cameras offer the potential to overcome the limitations of applications with standard cameras, such as low frame rate, high latency, low dynamic range, and high power consumption.
In fact, because of these promising features, this emerging camera has attracted attention from the industry, and there have been several commercial products, such as {\it asynchronous time-based image sensor} (ATIS) \cite{posch2010qvga}, {\it dynamic vision sensor} (DVS) \cite{lichtsteiner2008128}, and {\it dynamic and active pixel vision sensor} (DAVIS) \cite{brandli2014240}.

In the event camera, each pixel responds to changes in its photocurrent $L = \log I$ (i.e., brightness). 
A stream of events $e_t = (t, x, y, p)$ is triggered at each pixel $(x, y)$ at time $t$ as soon as the intensity increases or decreases from the last event in the pixel.
Here, $p$ is the polarity of events, indicating the sign of the brightness change.
In other words, $p$ at time $t$ and at pixel $(x, y)$ can be written as 
\begin{equation}
    p = 
    \begin{cases}
    1 & (L(x, y, t) - L(x, y, t_{prev})) = C) \\
    -1& (L(x, y, t) - L(x, y, t_{prev})) = -C) \\
    \end{cases},
\end{equation}
where $C$ and $t_{prev}$ stand for a preset intensity threshold (positive number) and the time when the last event is triggered, respectively.

\subsection{Related Work}
To show the advantages of the event camera, we first discuss its applications in the recent robotics community.
Then, to situate the current work in the context of vision-based reinforcement learning with this novel sensor, we cover previous vision-based reinforcement-learning studies in robotics and explain how leveraging the event camera can potentially speed up the control of autonomous robots.

\subsubsection{Event Camera Applications}

Researchers have demonstrated applications utilizing the advantages of the event camera over standard cameras, such as low latency and high dynamic range.
They include basic computer-vision algorithms, such as object detection and tracking \cite{glover2016event}, and applied ones such as gesture recognition \cite{amir2017low} and video reconstruction \cite{rebecq2019eventstovideo}

Recently, studies have emerged in the robotics field utilizing the event cameras.
For example, Vidal et al. demonstrated simultaneous localization and mapping (SLAM) on quadcopters \cite{Vidal_2018}, and Falanga et al. also examined the difference in latency that affects high-speed control on drones between standard and event cameras \cite{falanga2019fast}.
Then, Dimitrova et al. showed low-latency control of quadrotors using event camera, enabling attitude tracking at speeds of over \ang{1600}/s \cite{dimitrova2019towards}.
Moreover, Delmerico et al. introduced large dataset taken by the event cameras on fast-moving drones \cite{delmerico2019we}.
For further application examples and algorithms, see \cite{gallego2019eventbased}.

As these examples illustrate, the event camera can potentially benefit the robotics community greatly, although studies combining the event cameras and robotics are few, and are only the beginning of its contribution.
Here, we anticipate our proposed approach can accelerate developing robots with equipped event cameras for various cases through reinforcement learning.

% However, not only recognition algorithms, but robotics applications utilizing the event camera are important and necessary to show its advantages, which would accelerate future research as well.

% These studies often require manually-designed controlling programs.
% Here, we anticipate the end-to-end learning approach such as reinforcement learning can augment the application possibilities.

% 認識アルゴリズムの開発にイベントカメラが使われるようになってきたが
% ロボティクスの文脈ではまだ少ない。
% なぜなら、・・・
% しかし実用を考えた時に、制御まで含めて現実世界でイベントカメラを使用することは大事。

\subsubsection{Vision-Based Reinforcement Learning in Robotics} % Riku

Many studies that apply reinforcement learning in robotics use camera images for their system's observation, as images generally provide rich information about surrounding environments.
This method is called vision-based reinforcement learning.
For example, Asada et al. demonstrated a robot that learned to shoot a ball into a goal using a standard TV camera attached to the robot \cite{asada1996purposive}.
With similar learning algorithms, they also demonstrated a robot that could to learn to collaborate with other robots in soccer games \cite{asada1999cooperative}.
In their cases, they classically encoded the image into several sub-states by analyzing the object's position in the image.

The recent development of deep neural networks has enabled us to handle high-dimensional data and thus, to apply reinforcement learning without such manually-encoding processes, i.e., end-to-end learning.
Deep Q-network (DQN) solved classic Atari 2600 games \cite{bellemare2013arcade} and realized human-level control through deep reinforcement learning \cite{mnih2015human}.
The output from the simulator was high-dimensional data ($210 \times 160$ video at 60 Hz with 128-color pallet), and was resized into an $84 \times 84$-dimensional input image.
This work triggered a large number of studies on deep reinforcement learning and the development of various techniques to improve learning processes.
As a result, the learning efficiency has risen \cite{hessel2018rainbow} and more complex tasks have become trainable, such as generating responses for conversational agents \cite{li2016deep}.

However, applying vision-based deep reinforcement learning in robotics is not a simple task.
As end-to-end reinforcement learning requires numerous trials for agents to reach the optimal policy, we are restricted by the inability to have robots perform actions over and over in the real world, which is too costly and involves safety problems \cite{gu2017deep}.
The literature has faced many trials to fill in the gap between the real and simulated environments.
For example, Andrei et al. proposed an effective method to utilize simulators for training models and then transfer them into real robots \cite{rusu2016sim}.
Their approach successfully demonstrated task learning from raw visual input on a fully actuated robot manipulator.
To date, many works have leveraged simulators to render target environments, train their models inside them, and transfer them into robots \cite{kahn2018self,zhang2017deep,james20163d}.

Although many applications have been proposed, almost all of them assume standard cameras as their input to reinforcement learning.
At this point, we are interested in whether replacing these cameras with the event cameras will also result in the success of reinforcement learning and thus in the faster control of robots in the real world.

\section{Proposed Method} % Riku
In this section, we describe how we achieved the reinforcement learning applications with event-based data input for robots.
First, we mention the general settings of reinforcement learning for the following discussion foundation.
Second, we formulate an image-like feature for reinforcement learning to emulate event data with comparison to other approaches that emulate a stream of events.
Finally, we explain the entire process of the learning, from defining a problem to launching a robot in the real world.
The implementation is open-sourced \footnote{https://github.com/EventVisionLibrary/momaku}.

\subsection{Reinforcement Learning}

Standard reinforcement-learning settings consider an agent exploring a given environment ($\mathcal{E}$) to achieve the desired task.
Through a sequence of observations, actions, and rewards, the agent interacts with the environment and learns the optimal policy.
Formally, the set of possible observations and actions are defined as $\mathcal{S}$ and $\mathcal{A}$, respectively.
The agent receives an observation $s_n \in \mathcal{S}$ and a reward $r_n$ from $\mathcal{E}$ at each step index $n$, and then takes the next action $a_{n+1} \in \mathcal{A}$ based on its policy.
At a given step index $n$ and $s_n$, the accumulated reward from the state can be written as $R_n = \sum_{k=0}^\infty \gamma^k r_{n+k}$, where $\gamma$ is a discount factor for later rewards.
The goal of reinforcement learning is to determine the optimal policy that maximizes $R_n$ at each step.

\subsection{Image-Like Feature of Events for Reinforcement Learning}

In this section, we explain how we formulate and emulate input features for reinforcement learning in a simulator, and also how we convert an actual stream of events into the features.

Some prior studies proposed simulators for the event camera.
To obtain accurate event data, Mueggler et al. rendered images in a 3D environment at a fixed high sampling frequency \cite{mueggler2017event}.
Then, Rebecq et al. proposed an adaptive rendering to produce a stream of reliable event timestamp data \cite{Rebecq18corl}.
However, because we leverage reinforcement learning setting and assume step-by-step formulation, it is not necessary to produce a stream of events.
Rather, if we obtain at least the accumulated event data between each step, say $n-1$ and $n$, we can use it as an observation $s_n$.
Hence, estimation of the event timestamp between frames, as proposed in past simulation methods, is not necessary in the step-by-step reinforcement learning.

Therefore, instead of emulating a stream of events, we take the difference between two successive frames with a certain threshold to create an image-like feature.
This method is more computationally efficient and is thus suitable for reinforcement learning simulator use, which usually requires a large number of action steps.
In fact, this approach is similar to what Kaiser et al. proposed \cite{kaiser2016towards}, although they did not apply reinforcement learning and mentioned it as future work.

Formally, if we denote the timestamp for step index $n$ as $t_n$, we assume $s_n$ is the accumulated event data from $t_{n-1}$ to $t_{n}$.
The pixel $(x, y)$ of $s_n$ can be written as
\begin{equation}
    s_n(x, y) = 
    \begin{cases}
    1 & (L(x, y, t_n) - L(x, y, t_{n-1})) \geq C) \\
    -1& (L(x, y, t_n) - L(x, y, t_{n-1})) \leq -C) \\
    0 & ({\rm otherwise})
    \end{cases}.
\end{equation}
Therefore, the observation $s_n$ shall be a $H \times W$-dimensional feature where $W$ and $H$ are the event camera's width and height, respectively.
In the implementation of the simulator, we use OpenGL\footnote{https://www.opengl.org/} to render an environment at each time $t_n$ to obtain $L(x, y, t_n)$.

At this point, the remaining problem is how to convert an actual event stream $e_t$ into this format $s_n$ when we transfer the agent to a real robot with an event camera attached.
This is achieved by abandoning the timestamp information from the event stream and processing it as an accumulated batch feature during the time interval between $t_{n-1}$ and $t_n$.
The procedure is given by Algorithm \ref{alg:convert-events}.

\begin{algorithm}[t]
\caption{Converting events into an image-like feature}
\begin{algorithmic}                  
\Require a stream of event ($e_t=(t, x, y, p)$), $t_n$, $W$, $H$
\State Initialize $s_n \leftarrow O_{H, W}$ 
\State Extract a subset of events $E \leftarrow \{e_\tau |t_{n-1} \leq \tau < t_n\}$
\State Ensure $E$ is sorted by the timestamp in ascending order
\For{$e_t \in E$}
    \State $t, x, y, p \leftarrow e_t$
    \State $s_n(x, y) \leftarrow p$
\EndFor
\end{algorithmic}
\label{alg:convert-events}
\end{algorithm}

\subsection{From Simulation to Real World}

\begin{figure}[t]
\centering
\includegraphics[width=0.5\textwidth]{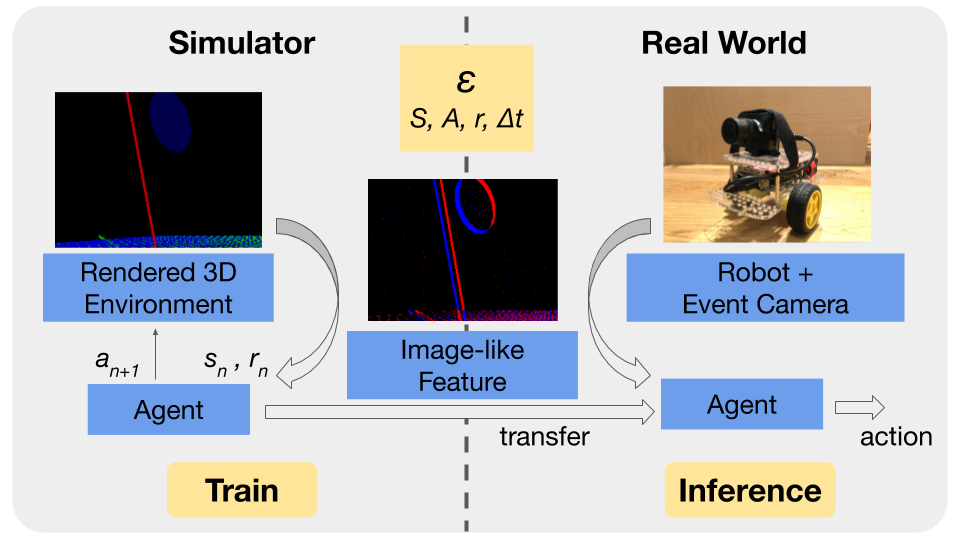}
\caption{Overview of our approach. In the learning process, an agent receives image-like feature $s_{n}$ and reward $r_{n}$, and then takes the next action $a_{n+1}$ in the simulator. In the deployment process, the learned agent can be deployed in a real-world robot to which an event camera is connected.}
\label{fig:overview}
\end{figure}
% https://docs.google.com/presentation/d/1nZvQ06QSLgabfWzRmrLSTei2ARkohDIZCsa1Fpu4WnA/edit#slide=id.g60e7ed640c_0_0

Figure~\ref{fig:overview} illustrates the entire training and deployment process.
First, we need to define an environment $\mathcal{E}$ where an agent will be trained along with observation space $\mathcal{S}$, action space $\mathcal{A}$, and reward setting.
We also need to configure the time interval $\Delta t$ between each step.
Thus, $t_{n+1} = t_{n} + \Delta t$ holds.
In standard vision-based reinforcement learning, the sampling rate of cameras provides a restriction for action frequency: $\Delta t \geq 0.033$~s for sampling at $30$~Hz or $\Delta t \geq 0.017$~s for $60$~Hz. %もし関連研究あれば
In contrast, we can lower this limit greatly by utilizing the event camera, which we explain in Section \ref{sec:experiment}.

Once above settings are configured, we employ a reinforcement algorithm to train a simulated agent.
An observation from the environment is provided by the method we mentioned in the previous section.
As we consider $s_n$ as an image-like feature, conventional reinforcement-learning algorithms can be applied.

Finally, the trained model can be transferred to a real robot with the event camera attached.
The interval between each action of the robot must be congruent with $\Delta t$.
The event stream $e_t$ is converted to $s_n$ by the interval $\Delta t$, and the trained model is applied to a sequence of input $s_n$.

\section{Experiment: Simulator Training}\label{sec:experiment}

\subsection{Problem settings}

We assumed the agent was a small car robot, simulating GoPiGo3 (Dexter Industries, Inc.)\footnote{https://www.dexterindustries.com/gopigo3/} equipped with DAVIS240 as a vision sensor \cite{brandli2014240}.
Hence, the simulated event camera had $240 \times 180$ pixels, which was located at the top of the agent.
We trained the agent for two types of simplified experiments: collision avoidance and object tracking.
Each of these tasks has been widely considered and explored as an important application for robotics.
For example, Michels et al. demonstrated a fast obstacle-avoidance algorithm using standard camera of 20~Hz \cite{michels2005high}.
However, even though the algorithm itself is fast, the control frequency was limited to a maximum of 20~Hz in their work.
In contrast, using our proposed procedure, the control frequency is no longer limited to the camera frame rate theoretically.
In our experiments, the step frequency $\Delta t$ was 0.01 (100~Hz) for both tasks.

\subsection{Simulator Environment}

We used various shapes of spheres and cubes as obstacles in the simulator.
The agent had actions of going forward, going backward, stopping, steering to right and left, some of which were enabled for each experiment.
To emulate the real camera, impulse noise was randomly added with the probability of occurrence at 0.001 independently for each pixel, convoluted into the image-like feature.

\subsection{Tasks}

\subsubsection{Collision Avoidance}

For the collision-avoidance task, we assumed a car running in a field where spheres randomly fell from the sky.
In each episode, one random sphere fell in front of the agent at a random moment, resulting in a collision if the agent continued running.
The agent had two actions of \{forward, stop\}.
Reward was defined as $r_{n} = -d_{n}^2 / 10.0$ for the Euclidean distance $d_{n}$ between the agent and the sphere falling in front of the agent.
Extra rewards were added by $0.2$ if the agent took the ``forward'' action.
If the agent collided with the sphere, the reward was $-50$.
The episode was done and reset when the agent collided with the sphere or acted with the maximum number of steps ($100$).

\subsubsection{Tracking}

As the tracking task, one sphere was thrown from an arbitrary point on a field, drawing parabola curve following the gravity. 
The agent's task was to follow the sphere by taking an action from three actions of \{forward, right, left\}.
Reward was set as $r_{n} = 10 (1 - \left| \theta_{n} \right|)$, where $\theta_{n}$ was the angle between the direction of the agent and that of the sphere seen from the agent.
Each episode was done and reset when the agent collided with the sphere or acted maximum number of steps ($100$).

\subsection{Reinforcement Learning Algorithm}

As one of the popular reinforcement learning algorithms, Double DQN with convolutional neural network was used to learn each task \cite{hasselt2015deep}.
The agent followed $\epsilon$-greedy exploration with $\epsilon = 0.1$.
For optimization, Adam with $\epsilon = 0.01$ was used \cite{kingma2014adam}.
Replay buffer of Double DQN was $10^6$ with $\gamma = 0.95$, and the target network update interval was 200 steps.
The neural network consisted of two convolutional layers and two fully connected layers, each of which followed by batch normalization and ReLU layers \cite{ioffe2015batch}.
The kernel size of the convolutional layers was three, their output channel sizes were two and four, and the number of dimensions of the fully connected layers was $100$.
The output dimension of the neural network was the number of actions for each task.
To train the neural network model, it took about one hour on GPU (GeForce GTX 1080, NVIDIA, Corp.).

\section{Result and Demonstration}

We first show the simulation results for the two experiments.
Next, we provide demonstration for the collision avoidance task on the real-world robot, a GoPiGo3 car equipped with the event camera.

\subsection{Simulation result}

\subsubsection{Collision avoidance}

\begin{figure}
\centering
\includegraphics[width=0.5\textwidth]{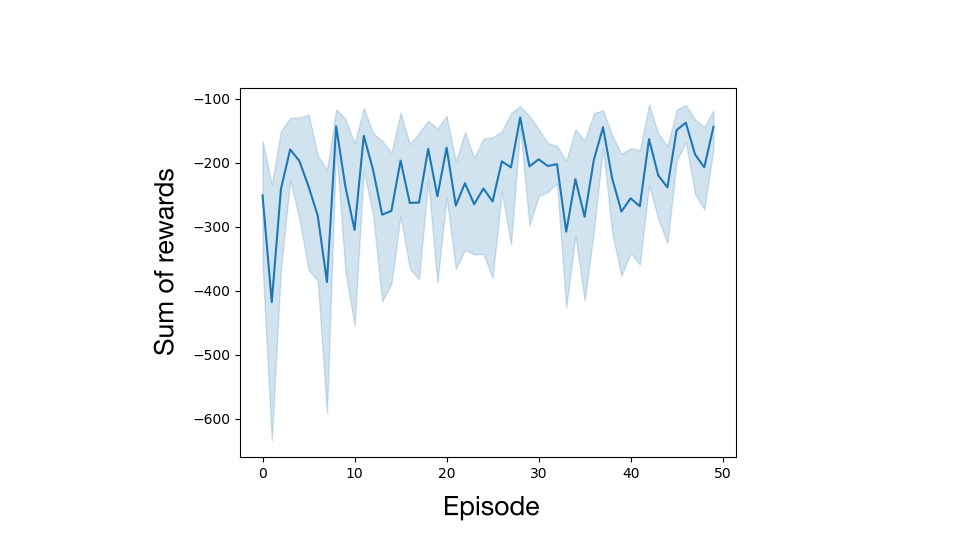}
\caption{Sum of rewards in each episode for collision-avoidance task. In each episode, the agent took at most 100 actions at 100 Hz and received rewards for each action. The mean and standard error over nine trials are described.}
\label{fig:result_avoidance}
\end{figure}

The sum of rewards, $R_{sum} = \sum_{k=0}^{100} r_{k}$, over each evaluation episode is shown in Fig. \ref{fig:result_avoidance}.
The result showed a confidence interval of nine experimental trials with different random seeds.
The sum of rewards increased along episodes, indicating that the deep neural network model successfully learned to avoid objects (stop) in front of the agent.

% format in https://docs.google.com/presentation/d/1nZvQ06QSLgabfWzRmrLSTei2ARkohDIZCsa1Fpu4WnA/edit#slide=id.g60e7ed640c_0_24

\subsubsection{Tracking}

\begin{figure}[h]
\centering
\includegraphics[width=0.5\textwidth]{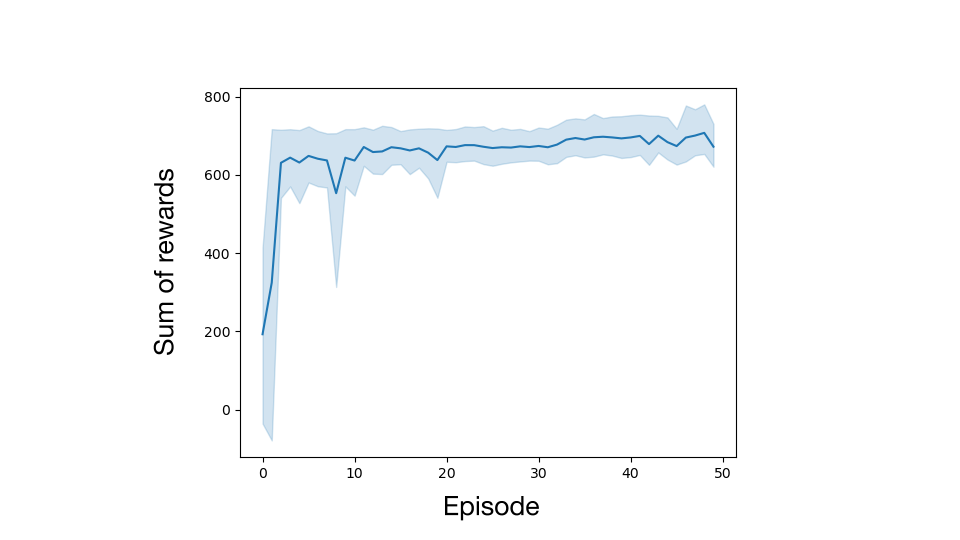}
\caption{Sum of rewards in each episode for tracking task. In each episode, the agent took at most 100 actions, and received rewards for each action. The mean and standard error over nine trials are described.}
\label{fig:result_tracking}
\end{figure}

The sum of rewards for the tracking experiment over each evaluation episode is shown in Fig. \ref{fig:result_tracking}.
The result showed confidence interval with nine experimental trials with different random seeds.
Similar to the avoidance experiment, it increased along episodes, showing that the neural network successfully learned how to track object and control itself toward the object in front.

\subsection{Demonstration}

\begin{figure}
\centering
\includegraphics[width=0.5\textwidth]{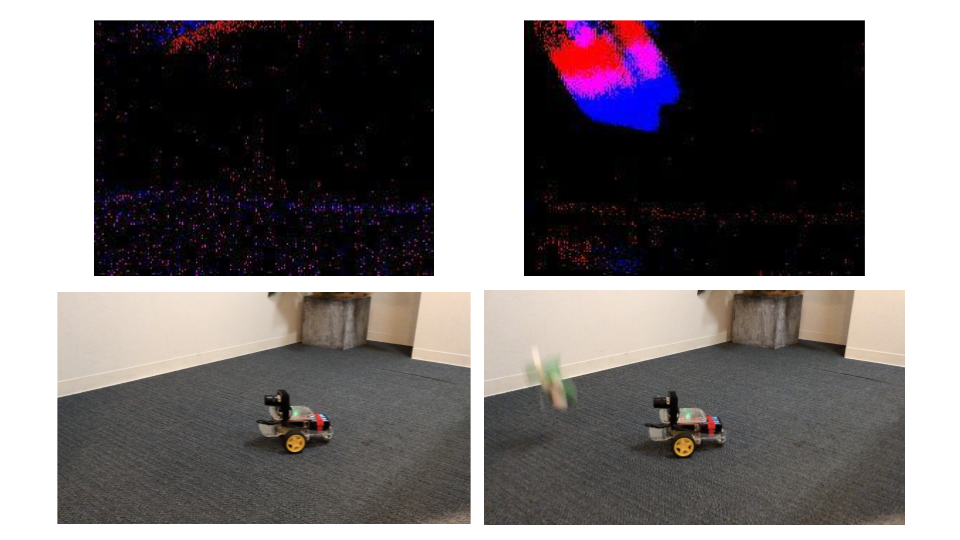}
\caption{Captures of the demonstration. GoPiGo3 equipped with DAVIS240 on it (lower images), and the reinforcement learning agent inferred the action from the image-like features (upper images). When an object was thrown in front of the agent, the car successfully stopped. For the detail, see the attached video.}
\label{fig:demo}
\end{figure}

After it was trained in the simulator for the collision-avoidance task, the agent was transferred and deployed into RaspberryPi 3 on GoPiGo3.
The setup image of the robot and the image-like features captured by DAVIS240 are shown in Fig. \ref{fig:demo}.
We used DAVIS240 as a sensor on the robot \cite{brandli2014240}.
To capture and handle the streaming data from the camera, we used the libraries of libcaer\footnote{https://gitlab.com/inivation/libcaer} and EventVisionLibrary\footnote{https://github.com/EventVisionLibrary/evl}.
The trained model ran on a server computer (MacBook Pro 2018, Apple Inc.), connected through WebSocket API and returning the inferred action to the robot.
The execution time of the inference for a single step was about 9~ms on the server.

The demonstration is shown in the video material \footnote{ https://youtu.be/xc2nKJ1TLTY}.

As a result, the car agent succeeded in stopping when an object was thrown in front of it suddenly.
It was demonstrated that our approach is effective to transfer and deploy the agent into real-world robots.

\section{Discussion and Conclusion}

We have presented the first application for reinforcement learning with the event camera.
After training in a simulated environment, the robot was able to perform desired movements in the real world, such as avoiding collisions and tracking an object.
Our robot was controlled faster at 100~Hz, compared to 30 or 60~Hz that are typical frequencies of standard cameras.
We believe that our approach opens a new possibility for fast and autonomous robots utilizing event cameras.

There remain some issues to be addressed to expand our result.
First, as we processed image-like features with high frequency and did not account for the timestamp information of events, future work shall integrate that information or process the stream of events asynchronously, which would provide more accurate information about the surroundings, thus making the control more accurate.
Second, we used WebSocket API to calculate the neural network inference on the server in our demonstration, since the robot car we used did not have enough computational resources.
Hence, for the next step, on-board inference without any network connection will be desired to implement to take the full advantage of the event camera.
In addition, our simulator and tasks were simple to demonstrate the feasibility of combining reinforcement learning with event camera for achieving faster control.
Still, the reinforcement-learning agent will be desired to handle more complex environment in order to tackle the real-world problems such as autonomous vehicles and drones today.
Therefore, it is expected to design more various simulation environments suitable for desired tasks and investigate the generalization ability of the agent.

\section*{Acknowledgments}

The authors would like to thank Hidenobu Matsuki for his support and guidance. This work was partially supported by MITOU Advanced funding program by the Ministry of Economics, Trade and Industry in Japan.

\newpage
\bibliographystyle{IEEEtran}
\bibliography{bibliography}

\end{document}